# An Asynchronous Federated Learning Platform for Lung Adenocarcinoma Grading via Foundation Models and Structural Pruning

**Moon-Il Joo [1], Md Ariful Islam Mozumder [1], Hee-Cheol Kim [1, 2 *]**

[1] Institute of Digital Anti-Aging Healthcare, Inje University, Republic of Korea

[2] College of AI & Computer Engineering, Inje University, Republic of Korea

* Corresponding author: heeki@inje.ac.kr

**Abstract:** Classification of lung adenocarcinoma into poorly differentiated (PD) and well-to-moderately differentiated (WMD) subtypes plays an important role in patient survival; however, consistent classification is difficult due to inter-observer variability among pathologists. Collecting data from multiple hospitals is also challenging due to patient privacy concerns. This study proposes a web-based federated learning system that shares only locally trained model weights without sharing patient data between institutions. The UNI2 foundation model and CLAM-SB were used to classify pathological whole slide images. A four-client federated learning system was constructed, and the NLST cohort was reserved exclusively for external validation. In centralized evaluation, joint training with CPTAC and TCGA outperformed single-institution training on external validation, confirming the necessity of multi-institutional data integration. FedAvg, FedProx, FedBN, and FedNova were compared as federated learning algorithms under identical experimental conditions. Test AUC ranked as FedBN, FedProx, FedNova, and FedAvg. FedNova showed the most balanced classification performance between PD and WMD. L1-norm structural pruning at 20%, 30%, and 40% sparsity was applied to reduce model transmission size. FedAvg, FedProx, and FedBN showed a tendency of decreasing Test AUC under pruning, while FedNova achieved improved Test AUC across all sparsity levels. A web-based application was developed supporting slide upload, local training, AI classification, and attention heatmap-based visualization. The combination of FedNova and 40% pruning was confirmed as a practical approach that improves communication efficiency and classification performance while protecting patient privacy.

## 1. INTRODUCTION

Lung cancer is one of the leading causes of cancer-related deaths worldwide, with an estimated 2.6 million new cases and 1.9 million deaths recorded globally in 2024 [1]. Among its subtypes, lung adenocarcinoma (LUAD) exhibits varying treatment options and prognosis depending on the degree of tissue differentiation. Accurate distinction between poorly differentiated (PD) and well-to-moderately differentiated (WMD) subtypes is a critical basis for determining the direction of patient treatment. The International Association for the Study of Lung Cancer (IASLC) proposed a grading system for invasive pulmonary adenocarcinoma in 2020, classifying tumors into three grades based on histological patterns [2]. However, this judgment is a subjective process in which pathologists directly examine tissue under a microscope, and results can differ depending on the reader even when viewing the same slide. The 2020 IASLC grading system showed discordant results in reproducibility assessment for grade 2 and 3 lesions, with an exact inter-observer agreement rate reported as low as 51.7% in some studies [3]. Against this background, interest has grown in systems that automatically analyze whole slide images (WSI) to classify the degree of differentiation [4].

Deep learning in pathological image analysis has been advancing rapidly [5]. In particular, multiple instance learning (MIL) divides large-scale WSIs into small patches for analysis and then classifies the entire slide using only slide-level labels, significantly reducing the annotation burden [6]. Attention-based MIL (ABMIL) was proposed to assign different attention weights to each patch, improving both classification performance and interpretability [7]. Building on this, clustering-constrained attention MIL (CLAM) further enabled data-efficient WSI classification through instance-level clustering [6]. More recently, Transformer-based MIL such as TransMIL has been proposed to capture long-range dependencies among patches, further advancing slide-level classification performance [8]. In addition, foundation models pre-trained on large-scale pathological images have recently emerged. UNI, a ViT-L/16-based model pre-trained on over 100,000 diagnostic WSIs across 20 tissue types, demonstrated strong generalization across diverse computational pathology tasks [9]. Studies combining such foundation models with MIL frameworks have reported further improvements in pathological classification performance [10]. There are several challenges in applying these technologies to real hospital environments. In particular, centralized approaches that aggregate all data in one place for training generally achieve high performance. In practice, however, securing data from multiple hospitals is difficult due to patient data security and privacy concerns [11]. A further problem is that scanners, tissue staining methods, and labeling criteria differ from hospital to hospital, causing models trained on data from one institution to perform significantly worse on data from another [12]. Domain shift across institutions has been identified as a major challenge in multi-site medical image analysis [13]. In this study, a centralized training model is used as the baseline to analyze the performance gap between single-institution and multi-institutional training, and federated learning (FL) is then applied for comparative evaluation. FL has attracted attention as a method to address data security and privacy concerns [14]. Each hospital trains a model locally without exporting its data, and only the model weights are transmitted to a central server to produce a global model, enabling collaborative learning across institutions without sharing raw data. However, existing FL studies face challenges including differences in data volume across hospitals and inconsistent labeling criteria for the same disease, resulting in label heterogeneity across institutions [15]. High communication costs incurred at every training round also present a practical barrier [16]. Systematic research on model compression techniques to reduce these costs remains insufficient [17]. Structural pruning based on L1-norm, which removes filters with low weight magnitudes, has been proposed as an effective approach to reduce model size while maintaining performance [18]. Applying such compression techniques in FL settings has shown promise for reducing communication overhead without significant accuracy degradation [19]. To address these issues, this study proposes a web-based asynchronous FL framework. The UNI2 foundation model and a CLAM-SB-based MIL framework are combined to classify PD/WMD from pathological image data. FedAvg, FedProx, FedNova, and FedBN algorithms are applied and compared, and L1-norm structural pruning is incorporated to reduce network communication overhead. An asynchronous architecture based on Flask and Celery is designed so that web application functions remain available without interruption during local model training and AI analysis. Attention heatmaps are also provided to enable visual verification of which tissue regions the AI model used as the basis for its decision. The main contributions of this paper are as follows.

- By setting centralized training as the baseline and quantitatively analyzing the generalization performance gap between single-institution and multi-institutional training, the necessity of multi-institutional data integration was experimentally confirmed.
- In the FL environment, FedAvg, FedProx, FedBN, and FedNova algorithms were combined with L1-norm structural pruning, and the performance of each combination was comparatively evaluated. A web browser-based FL platform was implemented to provide a system that pathologists can use with ease.
- By visually presenting the model's decision-making rationale through attention heatmaps, a foundation was established for intuitively interpreting AI results in clinical settings.

## 2. RELATED WORK

Deep learning-based grading classification research falls into two main directions: patch-level analysis and slide-level analysis. The patch-level approach is a supervised learning method in which WSIs are divided into patches of a fixed size and the tissue pattern of each patch is classified individually. High performance has been reported in classifying histological subtypes such as lepidic, acinar, papillary, micropapillary, and solid from resected LUAD slides divided into patches [20]. In addition, a CNN model trained to classify eight LUAD subtypes based on data labeled by consensus among 17 expert pathologists also achieved high performance [12]. Slide-level analysis is a weakly supervised approach that classifies entire slides using only slide-level labels, with MIL being the representative method. ABMIL, which assigns different attention weights to each patch, was proposed and improved slide classification performance [7]. CLAM, which combines attention-based learning with instance-level clustering, enabled data-efficient WSI classification using only slide-level labels [6]. More recently, TransMIL has been proposed to capture long-range dependencies among patches, further advancing slide-level classification performance [8]. However, the patch-level approach has the limitation of requiring high labeling costs to assign individual labels to each patch. Furthermore, slide-level analysis has largely been conducted on single-institution data, highlighting the need to evaluate performance using multi-institutional data.

Recently, foundation models pre-trained on hundreds of thousands of pathological images have been combined with MIL to achieve high classification performance. UNI [9] and its successor UNI2 were trained on 200 million image tiles extracted from more than 350,000 slides, demonstrating improved generalization performance across diverse histological classification tasks [9]. CONCH, a visual-language foundation model for computational pathology, has also demonstrated strong performance across multiple pathology tasks [10].

Self-supervised learning approaches have further enabled the development of powerful feature extractors without requiring labeled data [21].

Research on FL to address medical data privacy concerns has also been actively pursued. FedAvg, which exchanges only model weights without sharing raw data, is the most fundamental approach [14], but tends to show decreased performance when data distributions differ across clients. FedProx, which adds a proximal regularization term [15], FedBN, which retains batch normalization statistics locally [22], and FedNova, which corrects for differences in the number of local training steps [23], have been proposed as alternatives. Personalized FL approaches have also been explored to better handle client-specific data distributions [24]. The application scope has expanded to diverse cancer types, including meta-learning-based FL for LUAD diagnosis [25] and FL for prostate cancer Gleason grading [26]. However, when FL is applied to large-scale medical image data, substantial communication costs arise. Model pruning has been proposed as an effective technique to reduce the size of transmitted model updates [18], and combining pruning with FL has shown promise in reducing communication overhead [19]. Quantization-based approaches have also been explored for communication-efficient FL [27]. Despite these advances, systematic evaluation of compression techniques in the context of computational pathology FL remains insufficient [17]. This study proposes a web-based FL framework to address these issues and systematically analyzes the relationship between L1-norm structural pruning and classification performance across multiple FL algorithms.

## 3. MATERIALS AND METHODS

### *3.1. System Overview and Web Architecture*

As shown in Fig. 1, the system consists of a central server and four institutional clients (CPTAC-A, CPTAC-B, TCGA-A, TCGA-B). Each client transmits only locally trained model weights to the server, and the received weights are updated into a global model through an aggregation algorithm by the central server.

The client is designed as a web-based application. Accessible via browser alone without any separate software installation, this application is centered on pathologists and clinical researchers who lack specialized AI expertise. It provides whole slide image upload, real-time monitoring of local training and AI analysis progress, and visualization of PD/WMD prediction results through attention heatmaps. Local training and AI analysis based on foundation models require substantial computation time. This computational load poses a critical problem for web-based applications. Other functions cannot be used while local training and AI analysis are in progress. Therefore, an asynchronous task queue approach was adopted to decouple from the web interface. When local training and AI analysis are requested, they are performed independently in the background, while the remaining functions of the web application remain continuously available. This architecture enables uninterrupted use of the system in real-world deployment environments.

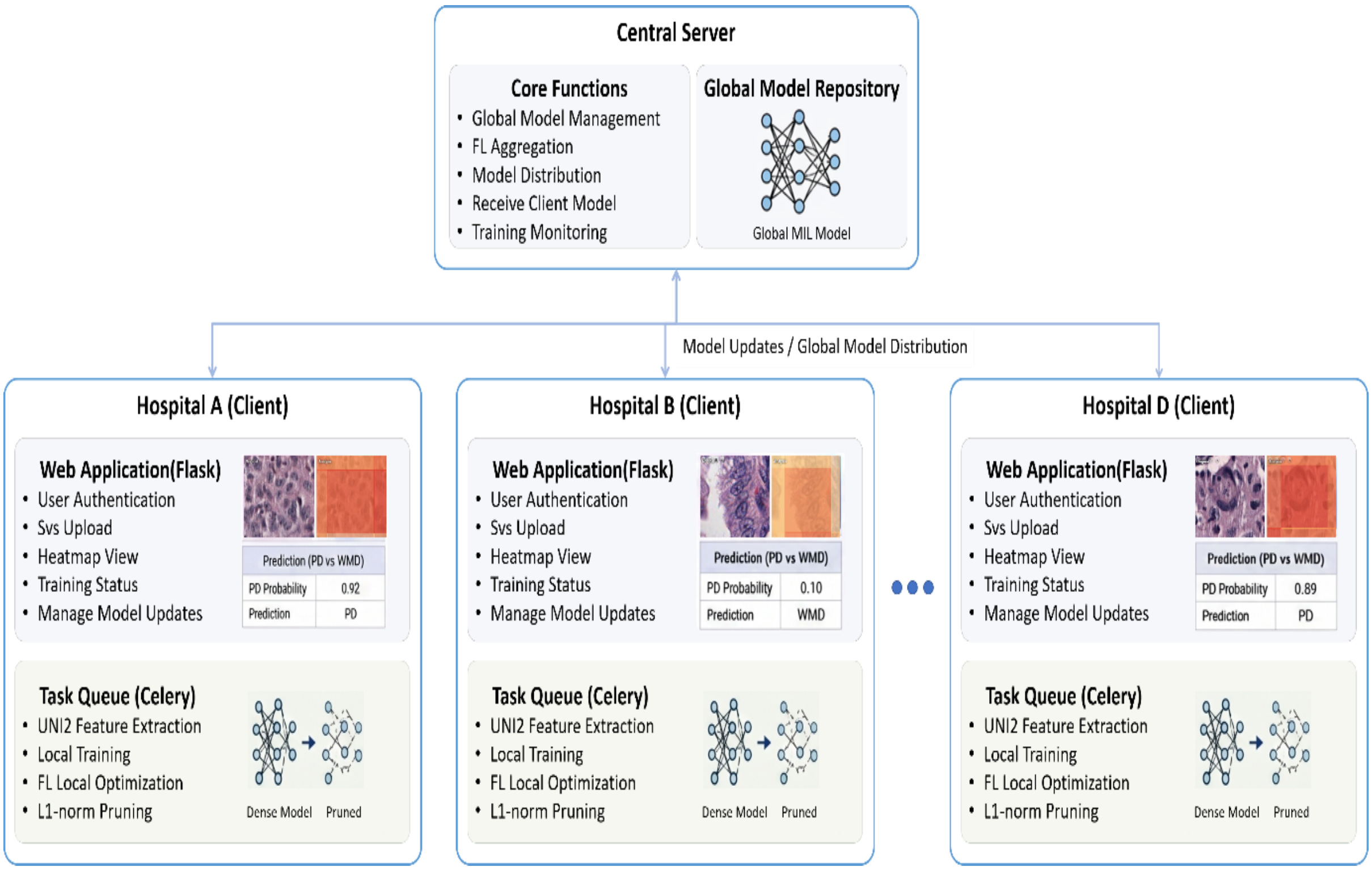

**Figure 1:** Overview of the web-based federated learning system architecture. The central server is responsible for aggregation, distribution, and repository of the global model, while each client handles Flask-based web application and Celery-based asynchronous local training and AI analysis.

### *3.2. Foundation Model-Based Multiple Instance Learning Framework*

Gigabyte-scale whole slide images (WSI) cannot be input in their entirety into an AI model due to GPU memory limitations. Therefore, this system employs a multiple instance learning (MIL) approach that divides slides into numerous small patch units for analysis. MIL extracts feature at the individual patch level, after which a MIL aggregator merges them into a single slide-level representation for final classification.

The feature extractor utilizes UNI2. UNI2 is a foundation model based on the ViT-H/14 architecture [28] trained on more than 350,000 pathological WSIs, capable of capturing subtle variations in cell nuclei and complex tissue arrangement patterns. WSIs are divided into non-overlapping patches of 256×256 pixels at 20× magnification, and a 1,536-dimensional embedding is extracted per patch. The resulting set of patch features (bag) is then used as input to the MIL aggregator.

PD/WMD classification is performed based on the CLAM-SB architecture. CLAM-SB is a MIL framework that utilizes attention-based pooling, automatically assigning high weights to regions that play a key role in classification among thousands of patches within the entire slide. Given a patch feature set $H = \{h_1, h_2, \dots, h_N\} \in R^{N \times D}$ (where feature dimension $D = 1536$), the final slide-level representation $z$ with attention weights applied is computed as follows.

$$ai = \frac{\exp\left(w^T\left(\tanh(Vh_i^T) \odot \sigma(Uh_i^T)\right)\right)}{\sum_{j=1}^{N}\exp\left(w^T\left(\tanh(Vh_j^T) \odot \sigma(Uh_j^T)\right)\right)} \quad (1)$$

$$z = \sum_{i=1}^{N} a_i h_i \quad (2)$$

Here, $V \in R^{L\times D}$, $U \in R^{L\times D}$ and $w \in R^{L}$ denote trainable projection parameters, and $\odot$ denotes the element-wise product. $a_i$ is the attention weight assigned to patch $i$. The aggregated slide representation $z$ is then passed through a binary classification head to ultimately predict whether the slide is PD or WMD.

The training loss function is constructed by weighted summation of the slide-level binary cross-entropy loss $\mathcal{L}_{bag}$ and the instance-level clustering regularization loss $\mathcal{L}_{inst}$ as follows.

$$\mathcal{L} = 0.7 \cdot \mathcal{L}_{bag} + 0.3 \cdot \mathcal{L}_{inst} \quad (3)$$

A bag weight of 0.7 was set to prioritize slide-level classification performance while suppressing overfitting through instance-level regularization. The Adam optimizer was used for optimization [29].

### *3.3. Communication-Efficient Federated Optimization*

There are three major barriers encountered when applying multi-institutional federated learning to real clinical environments. These are the non-IID problem in which data distributions differ across institutions, aggregation bias arising from differences in the number of slides, and domain shift caused by differences in scanners and staining methods. In addition, the communication burden of transmitting model updates of hundreds of MB per round also acts as a factor limiting practical applicability. To address these issues, this study applies and compares FedProx, FedNova, and FedBN. Furthermore, L1-norm structural pruning is combined to improve communication efficiency.

In FedProx, a client drift phenomenon occurs in federated learning where parameters gradually deviate from the global model as the client repeats local training over multiple epochs. This problem is suppressed by adding a proximal regularization term to the local objective function.

$$\mathcal{L}_i(w) = l_i(w) + \frac{\mu}{2}|w - w^t|^2 \quad (4)$$

The added term imposes a penalty to prevent the client parameter $\mathbf{w}$ from deviating excessively from the global model $\mathbf{w^t}$. Setting $\mu = 0$ makes it operate identically to FedAvg. As the value of $\mu$ increases, the degree of freedom in local updates decreases while convergence stability improves. In this study, $\mu \in \{10^{-4}, 5\times10^{-4}, 10^{-3}, 10^{-2}\}$ is used to determine the optimal value based on server validation AUC.

FedNova resolves aggregation bias arising from differences in the number of data and training steps across clients through normalization. The four clients in this study have different numbers of slides (CPTAC-A: 413, CPTAC-B: 430, TCGA-A: 290, TCGA-B: 320), resulting in up to a 1.48-fold difference in the number of local training steps $\tau_i$. The effective contribution of the CPTAC clients reaches approximately 58%, causing the global model to be biased toward specific institutions. FedNova resolves this imbalance through normalization that divides each client's update by the number of local steps $\tau_i$.

$$\delta_i = \frac{w^t - w_i^K}{\tau_i} \tag{5}$$

$$\tau_{eff} = \sum_{i=1}^{N} \frac{n_i}{N} \tau_i \approx 1119 \tag{6}$$

$$w^{t+1} = w^t - \tau_{eff} \cdot \sum_i \frac{n_i}{N} \delta_i \tag{7}$$

After normalization, the contributions of CPTAC and TCGA are balanced to 50% and 50%, respectively.

FedBN mitigates domain shift across institutions by separating normalization parameters from global aggregation. CPTAC data is captured with specific scanners while TCGA data is collected with various equipment from diverse institutions. As a result, even for the same tumor type, color distribution and tissue brightness differ across institutions. When the feature distributions of the input data itself differ in this way, averaging the statistics of the normalization layers across institutions can rather impair performance. A modified model (CLAM-SB-LN) that replaces the batch normalization of CLAM-SB with layer normalization is used. The scale parameter $\gamma$ and bias $\beta$ of the layer normalization are excluded from global aggregation and maintained locally at each institution.

$$w^{(t+1)} = \sum_i \frac{n_i}{N} w_i^K \setminus \Theta_{LN} \tag{8}$$

Parameters containing actual classification knowledge, such as the attention module and classification head, are shared across institutions, while the distributional statistics of input features are managed independently for each institution.

L1-norm structural pruning applies L1-norm based structural pruning before transmission. For each output neuron $j$ of each fully connected layer, the L1-norm of the weight change is computed.

$$s_j = |\Delta \mathrm{w_j}|_1 = \sum_k |\Delta w_{jk}| \tag{9}$$

Subsequently, the lower $\mathrm{s}\%$ of neurons with low importance are masked to zero to create a sparse vector, which is then computed as follows.

$$\widetilde{\delta_i} = \delta_i \odot \mathbb{1}[s_j \geq \tau_s] \tag{10}$$

### *3.4. Web-Based Asynchronous Federated Learning System Implementation and User Interface*

The web-based asynchronous federated learning system architecture was designed with a combination of FedNova and L1-norm 40% structural pruning. The client was designed as a web-based user application and an AI server. The web-based user application is a Flask-based web application that provides functions used in real time through the user interface, including login authentication, whole slide image (SVS) file upload, attention heatmap visualization, and real-time monitoring of local training and AI analysis progress. The AI server is built with asynchronous Celery and is dedicated to local training and AI analysis functions in the background. Celery is a system that performs tasks in the background separately from the web application and is essential for local training and AI analysis that require substantial computation time. This asynchronous architecture enables users to continue using the web application functions without interruption.

Communication between the server and clients is implemented via an HTTPS-based REST API. In each federated learning round, the client receives the global model from the server. Using this transmitted global model, the client performs local training, applies L1-norm 40% structural pruning to the trained local model, and transmits it to the server. The server performs FedNova aggregation on the received local models to update the global model. This updated global model is automatically transmitted to clients at the start of the next round.

Fig. 2(a) is a dashboard that monitors in real time the local training start time and progress status for each federated learning round. Fig. 2(b) shows the whole slide image viewer and attention heatmap analysis result screen. When a user uploads an SVS file, it is automatically transmitted to Celery to proceed with AI analysis. Upon completion, the analysis result screen displays the original slide image and the attention heatmap image side by side. On the right side of the screen, a navigator showing the enlarged area of the entire slide and a panel displaying the analysis results for the corresponding slide are provided.

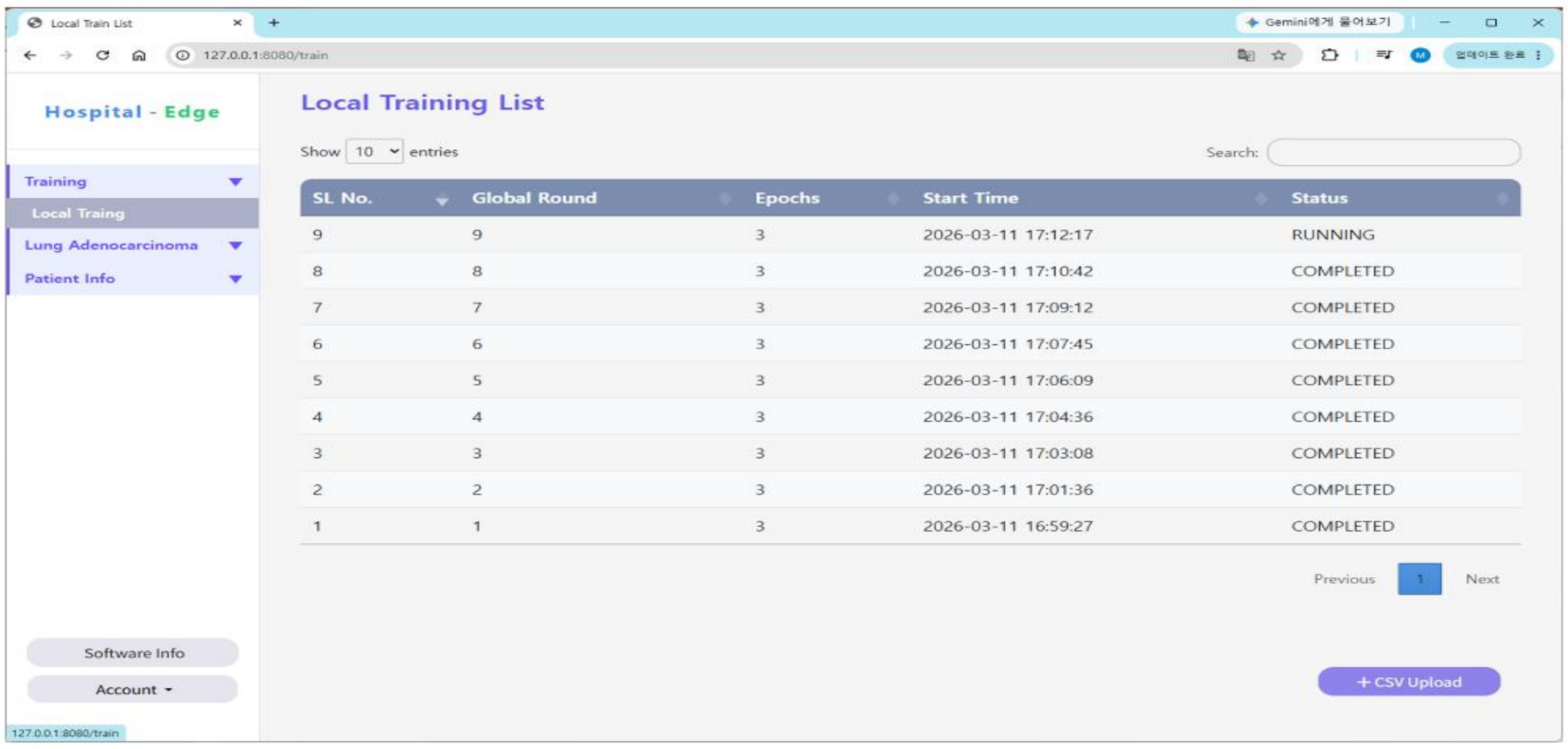

**(a) Local training status dashboard UI**

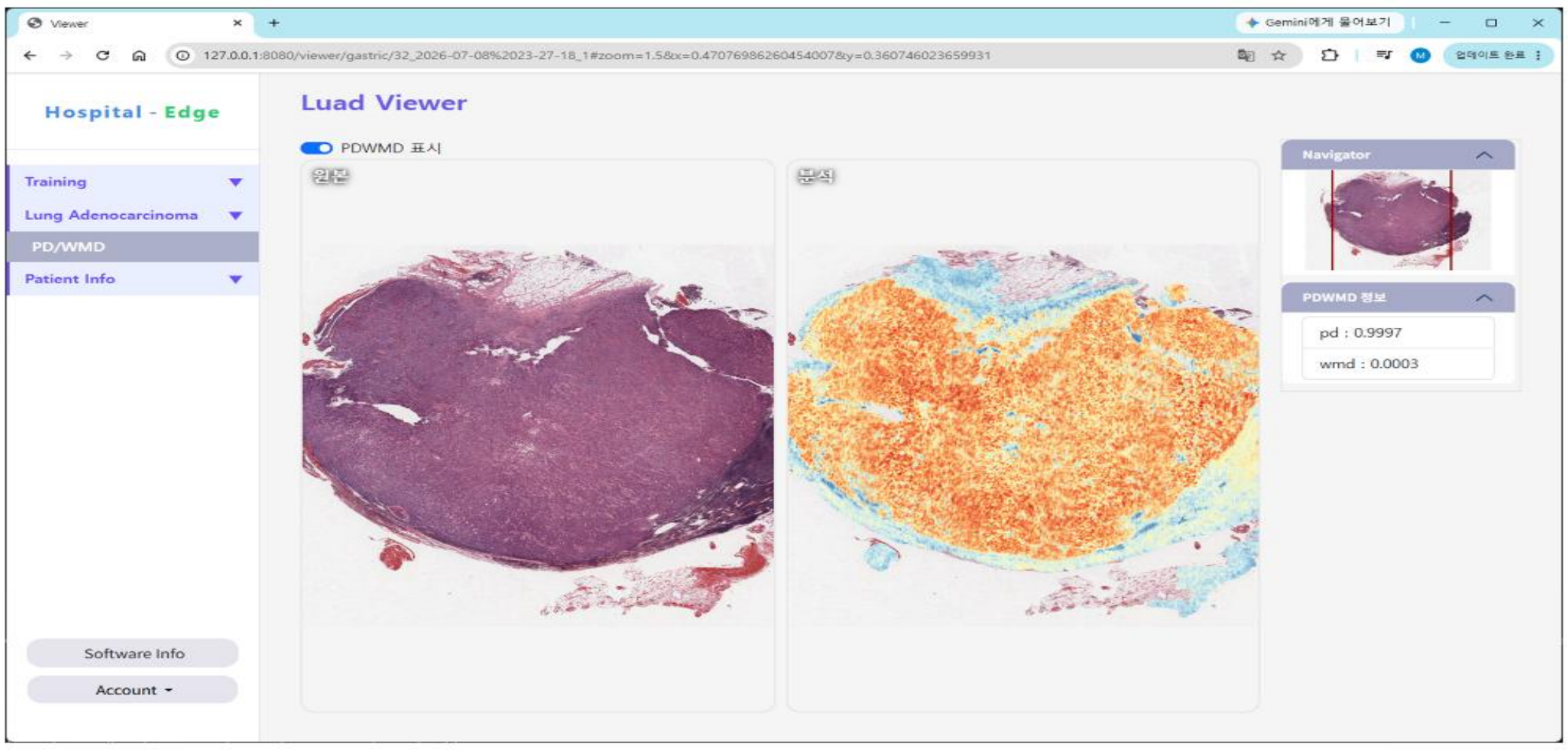

**(b) WSI and heatmap analysis result UI**

**Figure 2:** Web-based federated learning client interface. (a) Local training status dashboard. Displays in real time the training start time and progress status (RUNNING/COMPLETED) for each federated learning round (Global Round). (b) Whole slide image viewer and attention heatmap analysis result screen. The original H&E image (left) and attention heatmap (right) are displayed side by side, and the right panel provides a navigator showing the current magnified area within the entire slide and the PD/WMD prediction probabilities.

## *3.5. Datasets and Label Configuration*

This study utilizes three publicly available multi-institutional cohorts. CPTAC-LUAD and TCGA-LUAD are used for federated learning training and internal validation. NLST-LUAD is used for external independent validation. The composition of the three datasets is summarized in Table 1.

**Table 1:** Summary of the three datasets used in this study.

| **Dataset** | **Slides** | **Cases** | **PD** | **WMD** | **Label Criteria** | **Role** |
|---|---|---|---|---|---|---|
| CPTAC-LUAD | 1,003 | 214 | 292 | 711 | Tumor grade (G1–G2→WMD, G3–G4→PD) | Training/Validation |
| TCGA-LUAD | 728 | 249 | 272 | 456 | Histological subtype rule-based (Nature 2014) | Training/Validation |
| NLST-LUAD | 613 | 210 | 198 | 415 | Clinical de_grade field (de_type=8140) | External Validation |

CPTAC-LUAD consists of 1,003 slides (214 cases) collected from the Clinical Proteomic Tumor Analysis Consortium (CPTAC) [30]. Labels were assigned based on the tumor grade from pathology reports. G1 and G2 were classified as WMD, and G3 and G4 were classified as PD. The average number of slides per case is 4.69, the highest among the three cohorts.

TCGA-LUAD consists of 728 slides (249 cases) collected from The Cancer Genome Atlas (TCGA) [31]. It differs in character from CPTAC in that it is a multi-institutional cohort collected with various equipment from diverse institutions. Labels were assigned based on the predominant histological subtype rather than the pathology report grade. This criterion was derived using the annotations from Supplementary Table 1 of the Nature 2014 TCGA LUAD paper [32], classifying Lepidic, Acinar, and Papillary subtypes as WMD, and Solid and Micropapillary subtypes as PD. NLST-LUAD consists of 613 slides (210 cases) collected from the National Lung Screening Trial (NLST) [33] and is used for external independent validation in this study. Labels were assigned based on the de_grade field in the NLST clinical database, with de_grade 2–3 classified as WMD and de_grade 4–5 classified as PD. To target lung adenocarcinoma only, cases were selected by applying de_type = 8140.

## 3.6. Experimental Design

### *3.6.1. Centralized Data Split*

To objectively evaluate the effectiveness of federated learning, a baseline comparing performance when data is integrated and trained centrally is necessary. In this study, three centralized experiments were conducted as summarized in Table 2 to establish a comparison baseline for FL results.

**Table 2:** Centralized comparison experiment configuration.

| Experiment | Training Data | Validation | Test |
|---|---|---|---|
| A | CPTAC (10-fold CV) | CPTAC | NLST |
| B | CPTAC (10-fold CV) | CPTAC | TCGA |
| C | CPTAC+TCGA (10-fold CV) | CPTAC+TCGA | NLST |

Experiments A and B train a model on CPTAC and evaluate performance on NLST or TCGA, respectively, to verify external validation on different test sets. Experiment C trains a model by centrally combining CPTAC and TCGA and evaluates on NLST to verify external validation on a multi-institutional dataset. It also serves as the primary baseline for comparison with federated learning performance. All three experiments employed 10-fold cross-validation to ensure reliability [34].

### *3.6.2. Federated Learning Data Split*

CPTAC and TCGA were each split into two clients, resulting in a total of four clients. By splitting each dataset into two clients, it is possible to simulate a realistic multi-institutional environment where data heterogeneity exists even within the same cohort. The client-specific composition calculated based on actual CSV data is summarized in Table 3.

**Table 3:** Federated learning client configuration.

| Client | Cohort | Slides | Cases | PD | WMD |
|---|---|---|---|---|---|
| Client 1 | CPTAC-A | 413 | 91 | 129 | 284 |
| Client 2 | CPTAC-B | 430 | 90 | 142 | 288 |
| Client 3 | TCGA-A | 290 | 106 | 110 | 180 |
| Client 4 | TCGA-B | 320 | 105 | 119 | 201 |
| Server Val | CPTAC+TCGA | 278 | 71 | 98 | 180 |
| Test | NLST | 613 | 210 | 198 | 415 |
| Total | - | 1,453 | 392 | 500 | 953 |

### *3.7. Implementation Details*

The experimental infrastructure consists of a server and clients. The server is configured with an Intel Core i9-10900X (3.70 GHz) and four NVIDIA RTX 4090 GPUs (VRAM 24 GB each), running Ubuntu 22.04 and PyTorch 2.1 [35]. The server is responsible for global model aggregation, distribution, and validation in federated learning, as well as centralized model training and validation. The client is configured with a 13th-generation Intel Core i7-13700K (3.40 GHz) and an NVIDIA GeForce RTX 4080 (VRAM 16 GB), running Windows 11 and PyTorch 2.1. The client is responsible for local training and validation in federated learning and web application deployment. The detailed hyperparameters for preprocessing, model training, and federated learning implemented in this study are summarized in Table 4.

**Table 4:** Summary of Implementation Details and Hyperparameters.

| Category | Parameter / Configuration | Value & Search Space |
|---|---|---|
| Hardware | Server Environment | Ubuntu 22.04, i9-10900X, RTX 4090 (24GB) ×4 |
| | Client Environment | Windows 11, i7-13700K, RTX 4080 (16GB) |
| Preprocessing | Tissue Segmentation | Otsu (sthresh=8), Median Blur (mthresh=7) |
| | Artifact Filtering & Closing | white_thresh=15, black_thresh=50, close=4 |
| | Patch Extraction | 256×256 (20×), seg_level=−1, four_pt contour, padded |
| Model | Architecture | UNI2 (1,536-dim), CLAM-SB (small) |
| Optimization | Optimizer & Hyperparams | Adam (LR=$5\times10^{-5}$), L2 Decay=$1\times10^{-4}$, Batch=8 |
| | Loss Function | Slide level: Cross-entropy / Patch level: SVM |
| | Imbalance Handling | Weighted sampling enabled |
| Experiment | Centralized Baseline | 10-Fold Cross-Validation |
| | Federated Learning Setup | 15 Global rounds, 3 Local epochs |
| | Federated Algorithms | FedProx: $\mu$ ($10^{-4}$, $5\times10^{-4}$, $10^{-3}$, $10^{-2}$), FedNova, FedBN |
| | Pruning Sparsity | 20%, 30%, 40% |

### *3.8. Evaluation Metrics*

The primary metric is the Area Under the ROC Curve (AUC) [36]. Since the boundary of differentiation grade judgment can vary depending on the institution and reader even for the same slide, AUC, which is not dependent on a threshold, is adopted as the primary metric. Detailed classification metrics are derived from the confusion matrix. Based on a matrix consisting of four components (TP, TN, FP, FN), AUC, Sensitivity, Specificity, Accuracy, and Balanced Accuracy [37] are evaluated. The definition of each metric is provided in Table 5.

**Table 5:** Classification performance metrics derived from the confusion matrix.

| Metric | Formula | Description |
|---|---|---|
| AUC | Area under the ROC curve | Overall discriminative power independent of threshold (primary metric) |
| Sensitivity | TP / (TP+FN) | PD detection rate |
| Specificity | TN / (TN+FP) | WMD detection rate |
| Accuracy | (TP+TN) / Total | Overall accuracy |
| Balanced Accuracy | (Sensitivity+Specificity) / 2 | Class imbalance-corrected balanced accuracy |

In the federated learning experiments, two additional aspects are evaluated. The impact of the pruning module on classification performance and the actual communication reduction rate are assessed. The change in NLST AUC resulting from the use of the pruning module is computed for each algorithm to quantitatively analyze the impact of pruning. By analyzing the classification AUC and communication reduction rate, it is expected that a comparison between performance and efficiency can be made.

## 4. EXPERIMENTAL RESULTS

### *4.1. Centralized Comparison Experiment Results*

The validation results for Experiments A, B, and C are summarized in Table 6. The same evaluation metrics were applied to all three experiments, and each value is presented as the mean and standard deviation across 10 folds.

**Table 6**: Summary of centralized comparison experiment results.

| Experiment | AUC | Sensitivity | Specificity | Accuracy | Balanced Accuracy |
|---|---|---|---|---|---|
| A | 0.7228±0.0122 | 0.1859±0.0724 | 0.9571±0.0257 | 0.7080±0.0078 | 0.5715±0.0239 |
| B | 0.6424±0.0285 | 0.2434±0.1046 | 0.8640±0.0715 | 0.6321±0.0182 | 0.5537±0.0244 |
| C | 0.7705±0.0129 | 0.5975±0.1365 | 0.7677±0.1445 | 0.7127±0.0548 | 0.6826±0.0124 |

Experiment C achieved the highest discriminative power with an AUC of 0.7705±0.0129. Experiment A recorded an AUC of 0.7228±0.0122, while Experiment B recorded the lowest AUC of 0.6424±0.0285. The Balanced Accuracy of Experiments A and B remained at approximately 0.5, whereas Experiment C achieved the highest Balanced Accuracy of 0.6826±0.0124. In particular, the Sensitivity of Experiments A and B was only 0.1859±0.0724 and 0.2434±0.1046, respectively, indicating a severe class imbalance phenomenon. That is, models trained solely on CPTAC, as in Experiments A and B, showed a tendency to barely detect PD slides in external validation and predict most of them as WMD. However, Experiment C outperformed Experiments A and B across all evaluation metrics. This quantitatively supports the conclusion that the integration of multi-institutional data contributes substantively to classification performance.

### *4.2. Federated Learning Algorithm Comparison Results*

Federated learning was conducted using FedAvg, FedProx, FedNova, and FedBN with the same four-client configuration, 15 global rounds, and 3 local epochs. The model with the highest AUC among the 15 global rounds was tested on NLST. The classification threshold for each algorithm used both threshold=0.5 and the optimal threshold determined by Youden's Index [38]. Table 7 presents results obtained with threshold=0.5, and Table 8 presents results obtained with the optimal threshold. The classification results based on the optimal threshold for each algorithm are presented as confusion matrices in Fig. 3.

**Table 7:** Federated learning algorithm performance comparison (Threshold=0.5).

| **Algorithm** | **Threshold** | **Val AUC** | **Test AUC** | **Sensitivity** | **Specificity** | **Accuracy** | **Balanced Acc:** |
|---|---|---|---|---|---|---|---|
| FedAvg ($\mu$=0) | 0.5 | 0.8359 | 0.7589 | 0.6768 | 0.6795 | 0.6786 | 0.6782 |
| FedProx ($\mu=1\times10^{-4}$) | 0.5 | 0.8323 | 0.7723 | 0.5859 | 0.8048 | 0.7341 | 0.6953 |
| FedProx ($\mu=5\times10^{-4}$) | 0.5 | 0.8351 | 0.7713 | 0.5808 | 0.8048 | 0.7325 | 0.6928 |
| FedProx ($\mu=1\times10^{-3}$) | 0.5 | 0.8344 | 0.7690 | 0.5404 | 0.8241 | 0.7325 | 0.6823 |
| FedProx ($\mu=1\times10^{-2}$) | 0.5 | 0.8455 | 0.7626 | 0.5758 | 0.8120 | 0.7357 | 0.6939 |
| FedBN | 0.5 | 0.8160 | 0.7790 | 0.4444 | 0.8747 | 0.7357 | 0.6596 |
| FedNova | 0.5 | 0.8507 | 0.7706 | 0.6313 | 0.7663 | 0.7227 | 0.6988 |

**Table 8:** Federated learning algorithm performance comparison (Optimal Threshold).

| **Algorithm** | **Threshold** | **Val AUC** | **Test AUC** | **Sensitivity** | **Specificity** | **Accuracy** | **Balanced Acc:** |
|---|---|---|---|---|---|---|---|
| FedAvg ($\mu$=0) | 0.1622 | 0.8359 | 0.7589 | 0.8081 | 0.5831 | 0.6558 | 0.6956 |
| FedProx ($\mu=1\times10^{-4}$) | 0.4830 | 0.8323 | 0.7723 | 0.6010 | 0.8000 | 0.7357 | 0.7005 |
| FedProx ($\mu=5\times10^{-4}$) | 0.4886 | 0.8351 | 0.7713 | 0.6010 | 0.8000 | 0.7357 | 0.7005 |
| FedProx ($\mu=1\times10^{-3}$) | 0.4524 | 0.8344 | 0.7690 | 0.6010 | 0.8000 | 0.7357 | 0.7005 |
| FedProx ($\mu=1\times10^{-2}$) | 0.3908 | 0.8455 | 0.7626 | 0.6414 | 0.7711 | 0.7292 | 0.7063 |
| FedBN | 0.1783 | 0.8160 | 0.7790 | 0.6818 | 0.7398 | 0.7210 | 0.7108 |
| FedNova | 0.4940 | 0.8507 | 0.7706 | 0.6364 | 0.7614 | 0.7210 | 0.6989 |

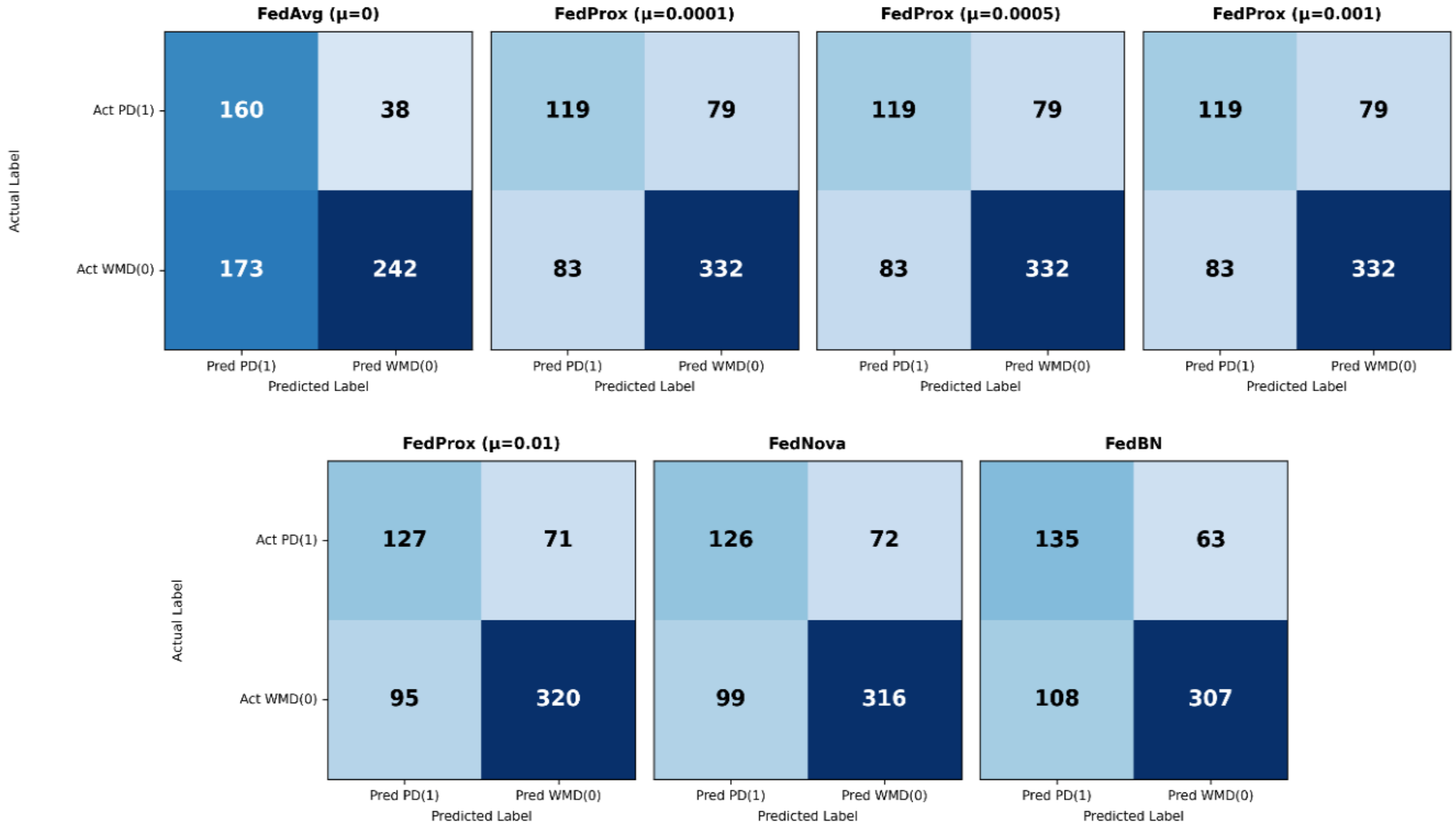

**Figure 3:** Confusion matrices of federated learning algorithms based on optimal threshold. The optimal threshold was determined as the value that maximizes Youden's Index on the validation set. Rows represent actual labels (Act PD (1), Act WMD (0)), and columns represent predicted labels (Pred PD (1), Pred WMD (0)).

In terms of Test AUC, FedBN achieved the highest value of 0.7790, while FedNova (0.7706) and FedProx (0.7723, 0.7713, 0.7690, 0.7626) outperformed FedAvg (0.7589). FedProx showed a tendency for Test AUC to decrease as μ increases. As shown in Table 7, in the performance evaluation with a uniform threshold of 0.5, FedNova demonstrated the most balanced performance. FedAvg and FedProx showed satisfactory performance. FedBN exhibited a high WMD detection rate with a Specificity of 0.8747 and a Sensitivity of only 0.4444. As shown in Table 8, performance differences according to threshold were observed when the optimal threshold was applied. FedBN required an extreme downward adjustment of the threshold to 0.1783 before Sensitivity improved to 0.6818, resulting in an improvement in Balanced Accuracy from 0.6596 to 0.7108. Similarly, FedAvg showed a comparable phenomenon with a threshold of 0.1622. When the optimal threshold is lowered in this manner, the model output probabilities are generally biased, making it difficult to use in actual clinical settings. In contrast, FedNova operates with a threshold of 0.494, which differs minimally from the default value, enabling stable data classification. This indicates that FedNova is the most suitable algorithm from the perspective of real-world clinical application.

### *4.3. Communication Efficiency Experiment Based on L1 Structural Pruning*

To reduce network transmission traffic occurring during the federated learning process, L1-norm based structural pruning was applied to the local model weights transmitted from clients to the server. In this study, the sparsity ratio was set at three levels of 20%, 30%, and 40%, and the optimal diagnostic threshold (Youden's Index threshold) was applied to evaluate classification performance. The performance changes by algorithm according to pruning intensity are shown in Table 9.

**Table 9:** Federated learning algorithm performance according to L1 structural pruning application.

| FL Algorithm | L1 0% Test AUC | L1 0% Bal. ACC | L1 20% Test AUC | L1 20% Bal. ACC | L1 30% Test AUC | L1 30% Bal. ACC | L1 40% Test AUC | L1 40% Bal. ACC |
|---|---|---|---|---|---|---|---|---|
| FedAvg (μ=0) | 0.7589 | 0.6956 | 0.6914 | 0.6490 | 0.6907 | 0.6504 | 0.6946 | 0.6485 |
| FedProx ($\mu=1\times10^{-4}$) | 0.7723 | 0.7005 | 0.6915 | 0.6479 | 0.6908 | 0.6504 | 0.6952 | 0.6482 |
| FedProx ($\mu=5\times10^{-4}$) | 0.7713 | 0.7005 | 0.6912 | 0.6480 | 0.6904 | 0.6516 | 0.6957 | 0.6468 |
| FedProx ($\mu=1\times10^{-3}$) | 0.7690 | 0.7005 | 0.6913 | 0.6490 | 0.6904 | 0.6504 | 0.6953 | 0.6478 |
| FedProx ($\mu=1\times10^{-2}$) | 0.7626 | 0.7063 | 0.6995 | 0.6485 | 0.6986 | 0.6462 | 0.7035 | 0.6540 |
| FedBN | 0.7790 | 0.7108 | 0.6893 | 0.6452 | 0.6849 | 0.6288 | 0.6919 | 0.6421 |
| FedNova | 0.7706 | 0.6989 | 0.7773 | 0.7107 | 0.7768 | 0.7094 | 0.7780 | 0.7189 |

There are differences in classification performance among FL algorithms. When L1 pruning at 20%, 30%, and 40% was applied, FedNova achieved Test AUC values of 0.7773, 0.7768, and 0.7780, respectively, all of which surpassed the L1 0% baseline. In particular, the best performance was achieved at L1 40%. However, FedBN and FedProx showed a decrease in Test AUC after all levels of pruning were applied. FedBN decreased by an average of 0.090, and FedProx decreased by an average of 0.074. In terms of compression efficiency, model size reductions of 23.7–40.4% relative to the original were achieved through L1 pruning. As shown in Table 10, applying L1 20% reduced the model size from the original 4.216–4.228 MB to 3.218 MB, with further reductions to 2.871–2.872 MB at 30% and 2.519–2.520 MB at 40%.

**Table 10:** Compression efficiency by L1 structural pruning level.

| Algorithm | Original (MB) | L1 20% (MB) | Reduction | L1 30% (MB) | Reduction | L1 40% (MB) | Reduction |
|---|---|---|---|---|---|---|---|
| FedAvg (μ=0) | 4.218 | 3.218 | 23.7% | 2.871 | 31.9% | 2.519 | 40.3% |
| FedProx ($\mu=1\times10^{-4}$) | 4.218 | 3.218 | 23.7% | 2.871 | 31.9% | 2.519 | 40.3% |
| FedProx ($\mu=5\times10^{-4}$) | 4.218 | 3.218 | 23.7% | 2.871 | 31.9% | 2.519 | 40.3% |
| FedProx ($\mu=1\times10^{-3}$) | 4.218 | 3.218 | 23.7% | 2.871 | 31.9% | 2.519 | 40.3% |
| FedProx ($\mu=1\times10^{-2}$) | 4.218 | 3.218 | 23.7% | 2.871 | 31.9% | 2.519 | 40.3% |
| FedBN | 4.216 | 3.218 | 23.7% | 2.872 | 31.9% | 2.520 | 40.2% |
| FedNova | 4.228 | 3.218 | 23.9% | 2.871 | 32.1% | 2.519 | 40.4% |

### *4.4. Visualization of Classification Basis Through Attention-Based Heatmaps*

In this study, patch-level attention weights were mapped to the original slide coordinates and visualized as heatmaps. The region of interest (ROI) was automatically detected centered on the high-contribution patches corresponding to the top 5% of attention scores among all patches, and a jet colormap was applied to express contribution levels from red (High) to blue (Low).

Fig. 4 presents the attention heatmap of a representative slide classified as PD (High-risk) together with the original H&E image. In the whole slide heatmap (b), high-attention regions were observed to be concentrated within the tumor tissue. In the ROI enlarged heatmap (d), high attention is concentrated in a concentric circle pattern centered at the central region, which corresponds to the area of solid growth pattern with high cell density and prominent nuclear pleomorphism confirmed in the ROI H&E (c). This suggests that the model utilizes tissue patterns in which normal glandular structures are lost and cellular atypia is intensified as the basis for PD classification. Fig. 5 presents the attention heatmap of a representative slide classified as WMD (Low-risk). In the whole slide heatmap (b), high-attention regions were broadly distributed across the entire tumor tissue, showing a contrasting pattern compared to the concentrated pattern toward a specific zone observed in the PD case. In the ROI H&E (c), well-preserved papillary and acinar glandular structures are clearly observed, and in the ROI heatmap (d), a pattern in which attention is distributed along these glandular structures in a vertical direction is confirmed. This indicates that the model utilizes the preservation of well-differentiated glandular structures as the basis for WMD classification.

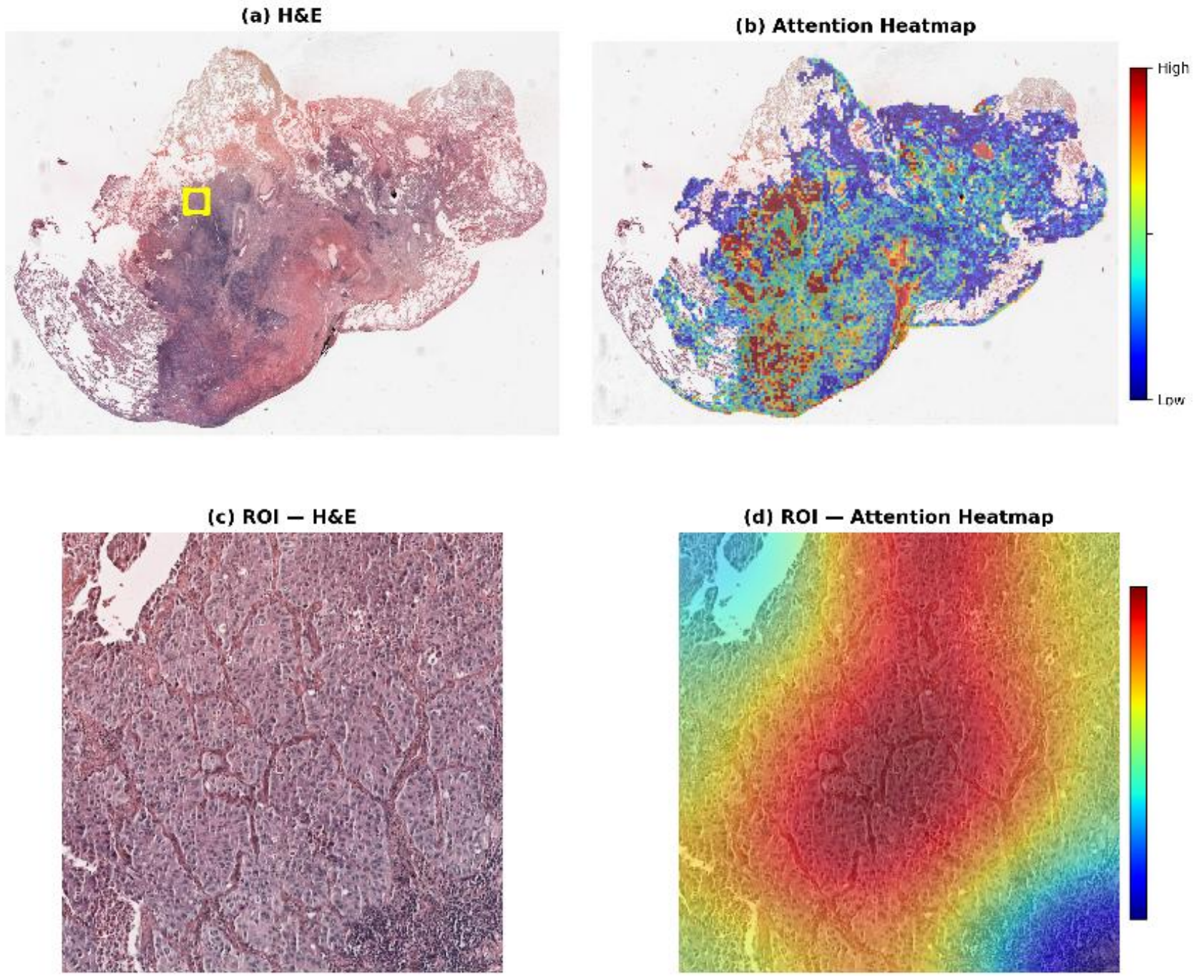

**Figure 4:** Attention-based heatmap visualization for PD. (a) H&E whole slide image (yellow rectangle: ROI location), (b) whole slide attention heatmap, (c) ROI H&E magnification, (d) ROI attention heatmap. The model's basis for PD classification is concentrated on regions of solid growth pattern with high cell density and prominent nuclear pleomorphism in the high-attention areas.

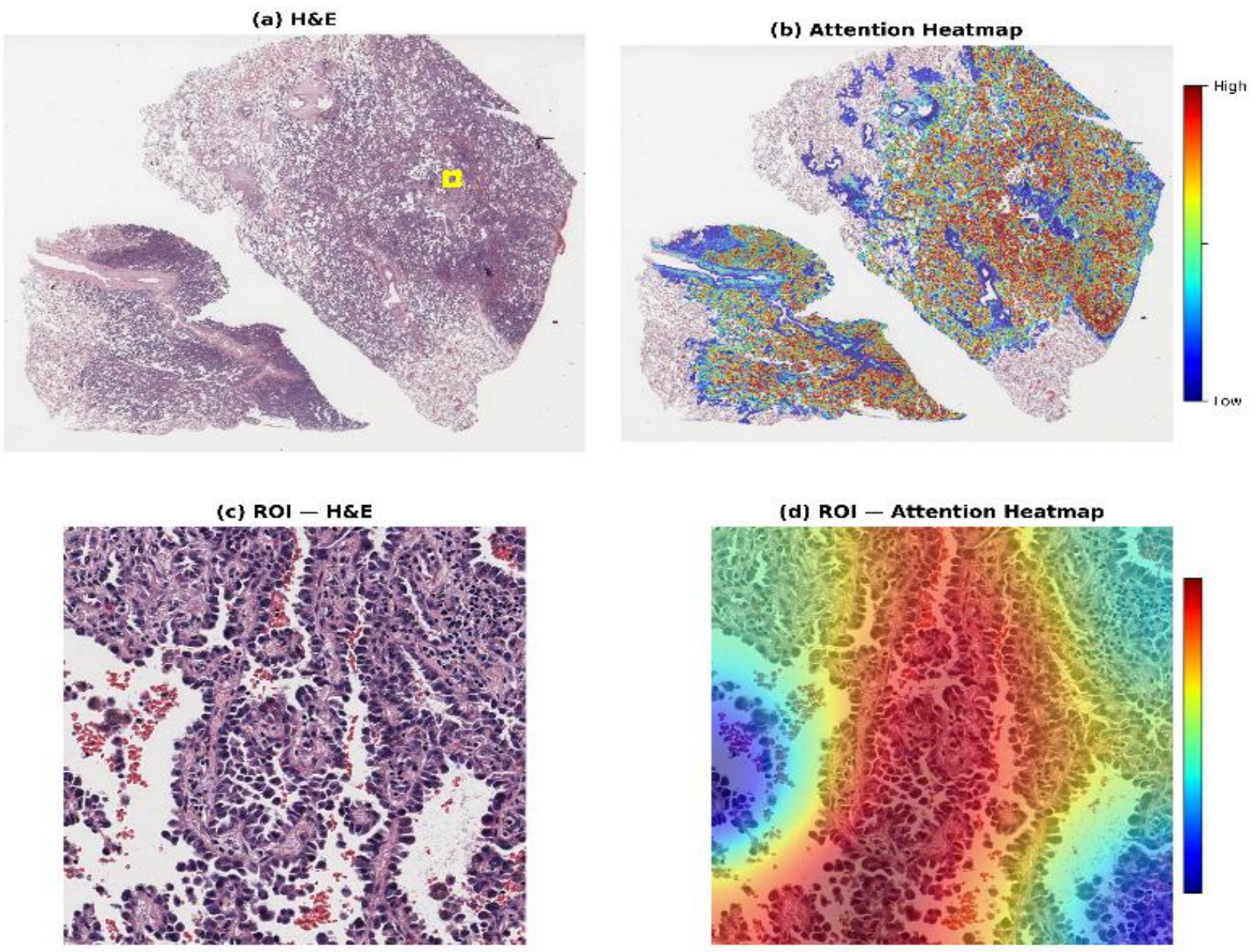

**Figure 5:** Attention-based heatmap visualization for WMD. (a) H&E whole slide image (yellow rectangle: ROI location), (b) whole slide attention heatmap, (c) ROI H&E magnification, (d) ROI attention heatmap. As the basis for WMD classification, high-attention regions are broadly distributed across the entire tumor tissue, and in the ROI, a pattern in which attention is distributed along papillary and acinar glandular structures is confirmed.

Comparing Fig. 4 and Fig. 5, a contrasting pattern is observed in which attention is concentrated in the cell-dense zone in PD slides, whereas it is distributed along glandular structures in WMD slides. This attention distribution appears to largely correspond to the histological features that pathologists focus on during visual examination, and can be utilized as supplementary evidence that enhances the interpretability of AI-based decisions.

## 5. DISCUSSION

This study performed PD/WMD differentiation grade classification of lung adenocarcinoma (LUAD) through web-based federated learning using multi-institutional data and quantitatively analyzed the effects of various FL algorithms and structural pruning on classification performance and communication efficiency. In the centralized experiments, a performance gap was observed between single-institution training with CPTAC (Experiments A and B) and integrated training with CPTAC+TCGA (Experiment C). In Experiments A and B, Sensitivity was only 0.1859 and 0.2434, respectively, indicating that these two experiments were nearly incapable of detecting PD slides. Experiment C demonstrated the highest performance with a Sensitivity of 0.5975 and an AUC of 0.7705. This implies that models trained on single-institution data perform poorly on external cohorts. This can be attributed to the model overfitting to staining methods and scanner characteristics during training, and suggests that using diverse datasets improves model classification performance.

Using the results of Experiment C as the baseline, the performance of federated learning was evaluated. Among the federated learning algorithms, FedBN recorded the highest Test AUC of 0.7790. However, the optimal threshold was as low as 0.1783, and even at threshold=0.5, Sensitivity did not reach half at 0.4444. Such a low threshold can be considered to indicate low stability for real-world system deployment. FedNova achieved an optimal threshold of 0.4940, close to the default value, enabling stable classification without threshold adjustment. Furthermore, balanced performance metrics such as Sensitivity of 0.6364 and Specificity of 0.7614 support this. FedProx achieved a good Test AUC of 0.7723 at $\mu=1\times10^{-4}$. Test AUC tends to decrease as $\mu$ increases, which appears to be due to the suppression of clients' ability to learn from their own data, thereby reducing the model's classification capability.

In the L1 structural pruning experiments, FedNova was the only algorithm that achieved an improvement in Test AUC compared to the unpruned baseline. Under the L1 40% condition, it demonstrated the best performance with a Test AUC of 0.7780 and a Balanced Accuracy of 0.7189. FedNova has a structure that mitigates data imbalance issues by correcting for differences in the number of local training steps across clients. The addition of L1 pruning appears to have further improved performance by removing unnecessary weights. In contrast, FedBN and FedProx showed Test AUC decreases of 0.059–0.081 after pruning. In particular, only marginal differences in Test AUC were observed across L1 20%, 30%, and 40%, suggesting that the Test AUC decrease is directly attributable to pruning.

In terms of communication efficiency, the combination of FedNova and L1 40% pruning achieved a reduction in local model size from 4.228 MB to 2.519 MB, a 40.4% decrease, while simultaneously improving classification performance. This result is particularly significant in that it demonstrates the feasibility of achieving communication compression without performance degradation in a FedNova environment, which is a core challenge of federated learning. This study has several limitations. First, the number of federated learning clients was limited to four, making it difficult to fully reproduce the diversity of real-world multi-institutional clinical environments. Second, the L1 structural pruning applied in this study was evaluated in the sparsity range of 20–40%; however, the performance changes at higher sparsity levels and the combined effect with float16 quantization remain to be explored in future research. Third, a reader study involving pathologists to validate the clinical utility of attention heatmaps was not conducted, which remains as a subject for future investigation.

## 6. CONCLUSION

This study proposed a web-based federated learning framework for PD/WMD differentiation grade classification of lung adenocarcinoma (LUAD) using diverse datasets. A system was implemented that enables multi-institutional collaborative learning to protect patient privacy and reduce network traffic. In particular, through the combination of FedNova and L1 40% structural pruning, the model size was reduced while classification performance was simultaneously improved. This is an experimentally demonstrated result showing that communication efficiency and classification performance can be improved at the same time. Furthermore, by visually providing the model's classification basis through attention heatmaps, a system was implemented that allows users in clinical settings to intuitively analyze results through visualization.

The practical advantages provided by this study are as follows. First, through asynchronous processing, tasks required for local model training and AI analysis are performed in the background, allowing other functions of the system to be used without interruption. Second, patient data is processed at each institution and is not transmitted externally, placing emphasis on personal information protection. Third, the combination of FedNova and L1 40% pruning reduces communication volume while improving classification performance, enabling stable system operation even in hospital environments with limited network bandwidth. In future research, we plan to expand the scope of validation to more institutions and diverse scanner environments, and to explore additional communication efficiency improvements through combination with float16 quantization.

- **Authorship contribution statement:** Moon-Il Joo; Writing – review & editing, Writing – original draft, Visualization, Validation, Supervision, Software, Methodology, Investigation, Formal analysis, Data curation, Conceptualization. Md Ariful Islam Mozumder; Writing – review & editing, Writing – original draft, Methodology, Investigation, Formal analysis. Hee-Cheol Kim; Writing – review & editing, Funding acquisition, Supervision, Conceptualization.

- **Data Availability Statement:** The datasets used in this study are publicly available through The Cancer Imaging Archive (TCIA). CPTAC-LUAD is available at https://doi.org/10.7937/k9/tcia.2018.pat12tbs, TCGA-LUAD is available at https://doi.org/10.7937/K9/TCIA.2016.JGNIHEP5, and NLST-LUAD is available at https://doi.org/10.7937/TCIA.HMQ8-J677.

- **Ethics statement:** This study used only publicly available, de-identified datasets obtained from The Cancer Imaging Archive (TCIA). No patient-identifiable information was accessed. Institutional review board (IRB) approval not required.

- **Declaration of generative AI and AI-assisted technologies in the manuscript preparation process:** During the preparation of this work, the authors used Gemini in order to improve language and readability. After using this tool/service, the authors reviewed and edited the content as needed and take full responsibility for the content of the publication.

- **Declaration of competing interests:** The authors declare that they have no known competing financial interests or personal relationships that could have appeared to influence the work reported in this paper.

- **Acknowledgments:** This work was supported by the Institute of Information & Communications Technology Planning & Evaluation (IITP) - Innovative Human Resource Development for Local Intellectualization program grant funded by the Korea government (MSIT) (IITP-2026-RS-2024-00436773).